\documentclass[pdftex,12pt]{article}  

\usepackage[a4paper, left=20mm, right=20mm, top=25mm, bottom=30mm]{geometry}  
\usepackage{amsfonts}   
\usepackage{amsmath}
\usepackage{graphicx}
\usepackage{hyperref}
\usepackage{orcidlink}

\title{Sign Language Recognition using Parallel Bidirectional Reservoir Computing}

\author%
{Nitin Kumar Singh$^{1}$\,\orcidlink{0000-0001-6368-8437},
Arie Rachmad Syulistyo$^{1}$\,\orcidlink{0000-0002-5933-1168},
Yuichiro Tanaka$^{1,2}$\,\orcidlink{0000-0001-6974-070X},\\
Hakaru Tamukoh$^{1,2}$\,\orcidlink{0000-0002-3669-1371}}

\date{%
$^{1}$Graduate School of Life Science and Systems Engineering,\\
Kyushu Institute of Technology, 2-4 Hibikino, Wakamatsu, Kitakyushu, 808-0196, Japan\\
$^{2}$Research Center for Neuromorphic AI Hardware, Kyushu Institute of Technology,\\
2-4 Hibikino, Wakamatsu, Kitakyushu, 808-0196, Japan\\[1ex]
{\normalfont singh.nitin-kumar991@mail.kyutech.jp, nitinmjpruiitp@gmail.com, syulistyo.arie-rachmad967@mail.kyutech.jp,tanaka-yuichiro@brain.kyutech.ac.jp, tamukoh@brain.kyutech.ac.jp\\
\vspace{0.5em}
\textcopyright\ 2026 IEICE,
 This work has been accepted by Nonlinear Theory and Its Applications (NOLTA), IEICE:Vol.E17-N,No.1,pp.-,Jan. 2026 }
}

\begin{document}

\maketitle


\begin{abstract} 
Sign language recognition (SLR) facilitates communication between deaf and hearing communities. Deep learning based SLR models are commonly used but require extensive computational resources, making them unsuitable for deployment on edge devices. To address these limitations, we propose a lightweight SLR system that combines parallel bidirectional reservoir computing (PBRC) with MediaPipe. MediaPipe enables real-time hand tracking and precise extraction of hand joint coordinates, which serve as input features for the PBRC  architecture. The proposed PBRC architecture consists of two echo state network (ESN) based bidirectional reservoir computing (BRC) modules arranged in parallel to capture temporal dependencies, thereby creating a rich feature representation for classification. We trained our PBRC-based SLR system on the Word-Level American Sign Language (WLASL) video dataset, achieving top-1, top-5, and top-10 accuracies of 60.85\%, 85.86\%, and 91.74\%, respectively. Training time was significantly reduced to 18.67 seconds due to the intrinsic properties of reservoir computing, compared to over 55 minutes for deep learning based methods such as Bi-GRU. This approach offers a lightweight, cost-effective solution for real-time SLR on edge devices.

\end{abstract}


\section{Introduction}
Sign language recognition enables effective communication between deaf and hearing-impaired people by interpreting hand movements, gestures, and facial expressions \cite{1}. SLR technology aims to automatically translate these gestures into spoken or written language, making communication more accessible and inclusive for everyone. The population of people with hearing impairments is continuously increasing, and this trend is expected to continue in the coming years. Consequently, SLR systems are essential in the current scenario, and researchers are seeking a cost-effective system that is accessible to the general public \cite{2}.

Deep learning-based models, such as convolutional neural networks (CNNs) and recurrent neural networks (RNNs), are widely used by researchers for developing SLR-based systems \cite{3}. However, due to several drawbacks, deep learning-based sign language recognition systems are unsuitable for edge devices, such as smartphones, tablets, or embedded systems \cite{4,5}. One of the primary challenges is the computational complexity of deep learning models, which necessitate substantial processing power to train and operate SLR systems. These models require high-end hardware, such as graphics processing units (GPUs) or tensor processing units (TPUs), and consume substantial memory and energy, which are often limited in edge devices. Furthermore, the training process for deep learning models is time-consuming and resource-intensive, making it impractical and costly for real-time SLR systems.

Santhalingam et al. describe the training process for various deep learning models used in American Sign Language (ASL) recognition. The authors discussed the 3D CNN models, which require approximately 20 hours to train on a GPU \cite{6}. 

Rastgo et al. provide a comprehensive review of advancements in sign language recognition systems over recent years, analyzing the impact of deep learning techniques such as CNNs and RNNs on recognition accuracy and highlighting the challenges posed by the need for extensive training \cite{7}.

Wong et al. discuss the use of deep learning techniques, such as hierarchical vision transformers, for SLR and raise concerns regarding their complexity and resource intensity \cite{8}.  

Starner et al. discuss the development and deployment of the PopSign ASL v1.0 dataset, along with the associated educational game designed to help hearing parents learn American Sign Language (ASL) through smartphone-based recognition. In this paper, the authors employ deep learning models, such as the long short-term memory (LSTM) model, for sign language recognition. However, as explained above, reliance on such complex models raises concerns about computational costs, which can be prohibitive for widespread deployment. The heavy dependence on computationally intensive models may hinder the development of scalable and cost-effective applications in sign language technology \cite{9}.  

Mohammadi et al. explore the development and comparison of spiking neural networks (SNNs) and traditional deep neural networks (DNNs) for static ASL gesture recognition. The authors implement four distinct SNN models on Intel’s neuromorphic platform, Loihi, and benchmark their performance against equivalent models deployed on the Intel Neural Compute Stick 2, a conventional edge-computing device. However, the use of deep learning–based algorithms and specialized hardware, such as Intel’s Loihi, can be considered expensive, particularly due to the specialized hardware requirements and the complexity involved in designing and deploying SNNs \cite{10}.  

Based on the above discussion, it can be concluded that deep learning based SLR systems still show multiple challenges in recognizing sign language effectively. Addressing these limitations is essential for developing efficient and cost-effective sign language recognition systems. In this paper, we propose an SLR system that combines MediaPipe (which extracts key features efficiently) with an echo state network (ESN)–based parallel bidirectional reservoir computing (PBRC) architecture, which is suitable for edge devices due to the core characteristics (lightweight nature) of reservoir computing.  

Reservoir computing (RC) is well-suited for resource-constrained environments because it requires training only the output layer, significantly reducing computational load and training time. Its internal reservoir—comprising fixed, randomly connected recurrent units—enables the system to capture temporal dependencies and nonlinear patterns in sequential data efficiently. Additionally, the simplicity and low power demands of RC make it competitive for real-time applications on embedded or edge devices. MediaPipe efficiently extracts key features from the video dataset, which are then fed into a PBRC-based architecture for gesture classification. The proposed PBRC-based framework is developed for isolated sign language recognition, where each sign (word-level unit) is recognized independently.

\section{Material and methods}
In this article, we utilized the WLASL100 video dataset, MediaPipe, and an echo-state network-based reservoir computing, which are described in detail below.

\subsection{Data collection and description}

\begin{figure}[t]
\centering
\includegraphics{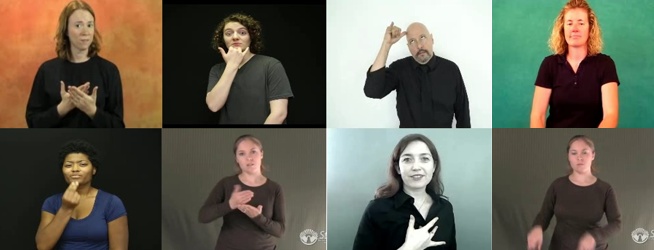}
\caption{Various signers performing different signs}
\label{fig:1}
\end{figure}

We used WLASL100 dataset for SLR \cite{11}.The WLASL100 dataset is a part of the Word-Level American Sign Language (WLASL) dataset, which is commonly used in research on SLR. WLASL100 includes video clips performed by multiple signers, which provide variations in hand movements, speed, and execution styles for different labels, as shown in Fig. \ref{fig:1}.

The WLASL100 dataset contains 100 unique labels, each representing a distinct American Sign Language word. The dataset is structured with 1,780 training videos, 258 validation videos, and 258 testing videos, ensuring a well-balanced distribution for model training and evaluation.

\subsection{MediaPipe}
MediaPipe is an open-source framework developed by Google that includes pre-built solutions for hand tracking, pose estimation, and face detection, making it useful in gesture-based recognition systems \cite{12}. In the context of sign language recognition, MediaPipe can easily extract meaningful landmarks and key points from video input. Its hand tracking module can identify 21 landmarks per hand, while the pose estimation module detects 33 body landmarks. These spatial points capture essential details about hand shapes, finger positions, and body posture, which are fundamental components of ASL.

MediaPipe's hand, face, and pose detection modules can be used to isolate critical visual features from each frame. These modules detect and track joints, fingertips, facial landmarks, and body poses in real-time, converting raw video data into structured, machine-readable features.

\subsection{Reservoir computing }

RC offers a unique approach to processing data, particularly when recognizing patterns and managing time series data. Unlike traditional neural networks, which undergo extensive training of the entire architecture, RC focuses on training only the output layer. This approach simplifies the training process and significantly reduces computational costs.

In RC, inputs are fed into the reservoir, where they are transformed. The reservoir has a fixed structure, but its internal states change dynamically with each new input, transforming the data into a high-dimensional representation. This process captures the temporal dynamics essential for tasks such as sign language recognition and speech recognition \cite{13,14}.

In this work, we use standard ESN-based reservoir computing (single uni-directional RC), BRC, and PBRC, as explained in detail below:

\subsubsection{Echo state network}

\begin{figure}[t]
\centering
\includegraphics{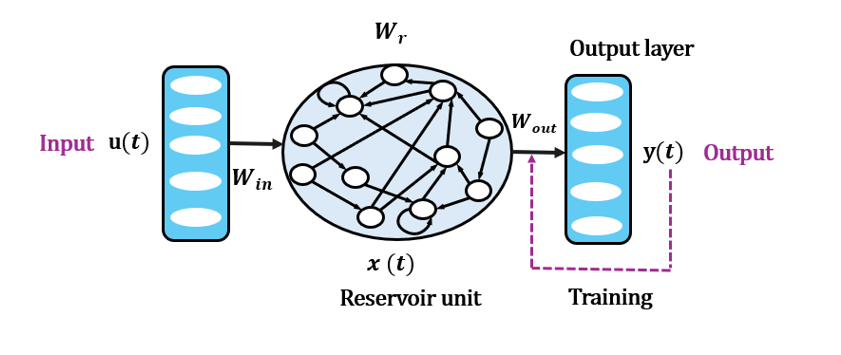}
\caption{Basic principle of operation of ESN-based RC}
\label{fig:2}
\end{figure}

In this work, all types of RC-based configurations maintain the properties of the echo state network.  
ESN is a type of reservoir computing that relies on the dynamics of a randomly connected recurrent neural network to process temporal data, as shown in Fig.~\ref{fig:2}.  

ESN-based RC is typically employed for time-series prediction, signal processing, and other tasks that require modeling sequential dependencies. ESNs are computationally efficient and relatively easy to implement \cite{15}.

In an ESN, the state update equation defines how the internal reservoir states evolve dynamically in response to incoming inputs, thereby creating a high-dimensional temporal representation. The output equation then applies a linear readout to these evolving states, mapping them into the desired output space for tasks such as classification or regression.
The state update equation, Eq.~(\ref{eq:state_update}), and output equation, Eq.~(\ref{eq:output}), of the ESN are given as follows \cite{16}:

\begin{equation}
x(t+1) = (1 - \alpha) x(t) + \alpha f\left(W_{\text{in}} u(t) + W_r x(t)\right)
\label{eq:state_update}
\end{equation}

\noindent
where \( x(t) \in \mathbb{R}^{N_r} \) is the reservoir state vector at time \( t \) and \( N_r \) is the number of reservoir nodes (with subscript \( r \) indicating ``reservoir''); \( u(t) \in \mathbb{R}^{N_{\text{in}}} \) is the input vector, \( W_{\text{in}} \in \mathbb{R}^{N_r \times N_{\text{in}}} \) is the input weight matrix that projects the input into the reservoir, and \( W_r \in \mathbb{R}^{N_r \times N_r} \) is the internal, fixed recurrent weight matrix of the reservoir. 

The function \( f(\cdot) \) is a nonlinear activation function, commonly chosen as \( \tanh(\cdot) \) or \( \sigma(\cdot) = \frac{1}{1 + e^{-x}} \), which introduces nonlinearity into the reservoir state dynamics. The parameter \( \alpha \in (0, 1] \) is the leak rate, which controls the speed of the reservoir states (smaller values of \( \alpha \) result in slower state updates, whereas values closer to 1 lead to faster updates).

\noindent
If \( \alpha = 1 \), the update equation simplifies to Eq.~(\ref{eq:simple_state_update}):

\begin{equation}
x(t+1) = f\left(W_{\text{in}} u(t) + W_r x(t)\right)
\label{eq:simple_state_update}
\end{equation}

\noindent
The output of the ESN is given by:

\begin{equation}
y(t) = W_{\text{out}} x(t)
\label{eq:output}
\end{equation}

\noindent
Here, \( y(t) \in \mathbb{R}^{N_{\text{out}}} \) is the output vector at time \( t \), and \( W_{\text{out}} \in \mathbb{R}^{N_{\text{out}} \times N_r} \) is the readout weight matrix, typically trained using ridge regression or another linear regression technique.

The input and reservoir weight matrices (\( W_{\text{in}} \) and \( W_r \)) are initialized randomly. The spectral radius of the reservoir is scaled to ensure that the echo state property (ESP) is maintained. The spectral radius of the recurrent weight matrix \( W_r \), denoted as \( \rho(W_r) \), is defined as the largest absolute eigenvalue of \( W_r \), i.e., \( \rho(W_r) = \max_i |\lambda_i| \), where \( \lambda_i \) are the eigenvalues of \( W_r \). To ensure the ESP, the matrix \( W_r \) is normalized and rescaled using the formula \( W_r \leftarrow \rho \cdot \frac{W_r}{\rho_1(W_r)} \), where \( \rho \) is a predefined spectral radius hyperparameter, typically chosen in the range \( (0, 1] \). This scaling controls the temporal memory and stability of the reservoir: a larger \( \rho \) helps retain longer input history, while \( \rho_1 \) improves stability.

\subsubsection{Bidirectional reservoir computing}

In a standard ESN, inputs flow in one direction, whereas bidirectional reservoir computing (BRC) can process the input sequence in both forward and backward directions simultaneously \cite{17}.

\begin{figure}[t]
\centering
\includegraphics{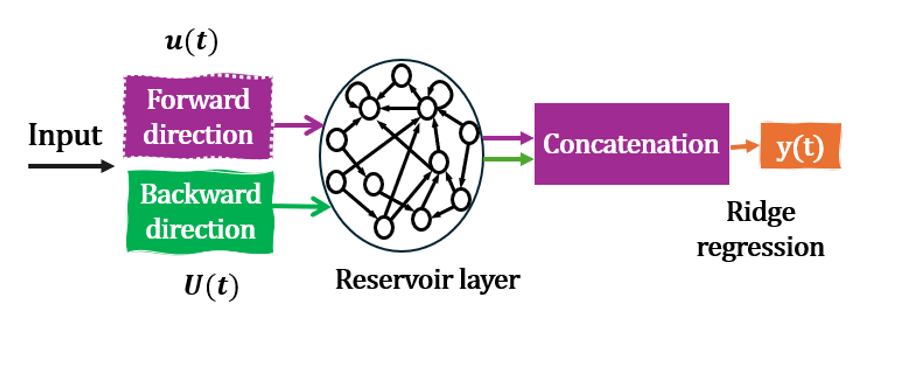}
\caption{Working principle of bidirectional reservoir computing-based architecture}
\label{fig:3}
\end{figure}

A bidirectional ESN maps its inputs to a high-dimensional state space, and the outputs from both reservoirs are concatenated to perform prediction or classification based on the complete temporal context of the input data, as shown in Fig.~\ref{fig:3}.

The state update equation for processing the data in the forward direction is given by Eq.~(\ref{eq:forward_state_update}) \cite{18}.
\begin{equation}
x_f(t+1) = (1 - \alpha) x_f(t) + \alpha \, \sigma(W_r x_f(t) + W_{\mathrm{in},f} u(t))
\label{eq:forward_state_update}
\end{equation}
where \( x_f(t) \) denotes the state vector of the forward direction at time \( t \), \( u(t) \) is the input vector, \( W_r \) is the fixed internal reservoir weight matrix shared during the processing of data in forward and backward directions, \( W_{\mathrm{in},f} \) is the input weight matrix for the forward direction, and \( \sigma \) is the activation function (e.g., \( \tanh \) or ReLU). In this study, we use \( \tanh \) as the activation function. The parameter \( \alpha \in (0, 1] \) is the leak rate that controls the update speed of the forward reservoir states. If \( \alpha = 1 \), the state is fully updated at each time step, completely replacing the previous state with the new computation.
The state update equation for the backward direction is similarly defined as Eq.~(\ref{eq:backward_state_update}).
\begin{equation}
x_b(t+1) = (1 - \alpha) x_b(t) + \alpha \, \sigma(W_r x_b(t) + W_{\mathrm{in},b} U(t))
\label{eq:backward_state_update}
\end{equation}
where \( x_b(t) \) is the state vector of the backward direction at time \( t \), processing the reversed input sequence \( U(t) \). The internal reservoir weight matrix \( W_r \) is shared during the forward direction as well, while \( W_{\mathrm{in},b} \) is the input weight matrix specific to the backward direction. The activation function \( \sigma \) is the same as in the forward direction. The parameter \( \alpha \in (0, 1] \) is the leak rate that governs the update speed of the backward reservoir states. 

In bidirectional reservoir computing, the input sequence is reversed by flipping its order in time while keeping all features unchanged. In Python, this is done using array slicing to reverse the time axis, as explained below:  
The reversed input sequence \( U(t) \) is obtained by indexing the input tensor \( X \) in reverse temporal order using array slicing, i.e., \( X[:, ::-1] \). The input tensor \( X \in \mathbb{R}^{N_U \times T} \), where \( N_U \) is the number of input features and \( T \) is the number of time steps in the sequence. This slicing operation reverses the temporal dimension while maintaining the input dimensionality. 

The forward and backward directions (reverse sequence of input) of BRC are then concatenated to give the final output, as expressed by Eq.~(\ref{eq:combined_output}).
\begin{equation}
y(t) = W_{\text{out}} (x_f(t) \oplus x_b(t)) 
\label{eq:combined_output}
\end{equation}
where \( y(t) \) is the output vector at time \( t \), \( W_{\text{out}} \) is the trained output weight matrix that maps the concatenated states from both the forward and backward reservoirs to the target output, and \( \oplus \) denotes the concatenation of both states of BRC. We used ridge regression to train \( W_{\text{out}} \) and classify the labels.

\subsubsection{Parallel bidirectional reservoir computing}

\begin{figure}[t]
\centering
\includegraphics{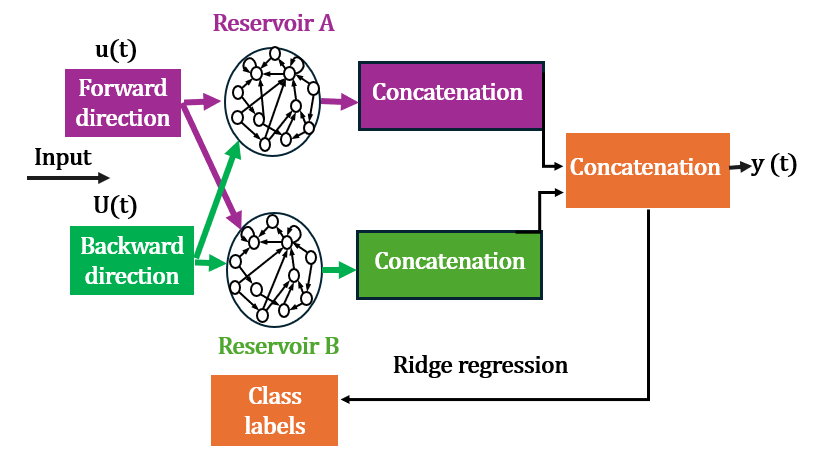}
\caption{Schematic diagram illustrating the working of parallel bidirectional reservoir computing}
\label{fig:4}
\end{figure}

The proposed reservoir architecture incorporates two echo state network reservoirs configured in parallel, each operating bidirectionally to process temporal features from the sign language video dataset. By reversing the input sequence for the backward pass, the architecture captures both past and future dependencies within each reservoir. The parallel configuration enhances accuracy by capturing comprehensive temporal dependencies and extracting diverse features from input sequences. Each reservoir's forward and backward states are concatenated, and the outputs of both reservoirs are further concatenated to create a rich feature representation for classification, as shown in Fig.~\ref{fig:4}.

The state update equations for the two bidirectional ESNs (Reservoir A and Reservoir B) are defined as follows.

Each reservoir processes the input sequence bidirectionally but shares the same internal recurrent weight matrix for both forward and backward directions, as shown in Fig.~\ref{fig:4}. For Reservoir A, the same matrix \( W^A \) is used for both directions; similarly, Reservoir B uses a single matrix \( W^B \).

PBRC comprises two independently configured BRC units, namely Reservoir A and Reservoir B. Each reservoir is composed of a forward and a backward processing stream to handle temporal information from both directions, given by Eqs.~(\ref{eq:forward_state_update_A_sharedW}--\ref{eq:final_concatenation_sharedW}). 

For Reservoir A, the forward state update at time \( t+1 \) is defined as:
\begin{equation}
x_f^A(t+1) = (1 - \alpha) x_f^A(t) + \alpha \, \tanh\left(W^A x_f^A(t) + W_{\text{in},f}^A u(t)\right),
\label{eq:forward_state_update_A_sharedW}
\end{equation}
and the backward state update is given by:
\begin{equation}
x_b^A(t+1) = (1 - \alpha) x_b^A(t) + \alpha \, \tanh\left(W^A x_b^A(t) + W_{\text{in},b}^A U(t)\right),
\label{eq:backward_state_update_A_sharedW}
\end{equation}
where \( W^A \) is the shared internal recurrent weight matrix, \( W_{\text{in},f}^A \) and \( W_{\text{in},b}^A \) are the input weight matrices for forward and backward directions, respectively. The vectors \( x_f^A(t) \) and \( x_b^A(t) \) represent the state vectors in the forward and backward directions, and \( U(t) \) denotes the reversed input. 

Similarly, for Reservoir B, the forward and backward state updates are given as:
\begin{equation}
x_f^B(t+1) = (1 - \alpha) x_f^B(t) + \alpha \, \tanh\left(W^B x_f^B(t) + W_{\text{in},f}^B u(t)\right),
\label{eq:forward_state_update_B_sharedW}
\end{equation}
\begin{equation}
x_b^B(t+1) = (1 - \alpha) x_b^B(t) + \alpha \, \tanh\left(W^B x_b^B(t) + W_{\text{in},b}^B U(t)\right),
\label{eq:backward_state_update_B_sharedW}
\end{equation}
where \( W^B \) is the shared internal recurrent weight matrix of Reservoir B, \( W_{\text{in},f}^B \) and \( W_{\text{in},b}^B \) are the input weight matrices for the forward and backward directions, respectively. The vectors \( x_f^B(t) \) and \( x_b^B(t) \) denote the state vectors corresponding to the forward and backward processing streams of Reservoir B, and \( U(t) \) denotes the reversed input sequence. 

The final concatenated state vector across both reservoirs is computed as:
\begin{equation}
x(t) = x^A(t) \oplus x^B(t) = \bigl(x_f^A(t) \oplus x_b^A(t)\bigr) \oplus \bigl(x_f^B(t) \oplus x_b^B(t)\bigr),
\label{eq:final_concatenation_sharedW}
\end{equation}
and the output is generated using the readout weight matrix \( W_{\text{out}} \) as shown in Eq.~(\ref{eq:parallel_bidirectional_output_sharedW}).
\begin{equation}
y(t) = W_{\text{out}} x(t),
\label{eq:parallel_bidirectional_output_sharedW}
\end{equation}
where \( W_{\text{out}} \) is trained using ridge regression to map the high-dimensional reservoir states to class labels.

\subsection{Ridge regression}

We employed ridge regression to train and fine-tune the PBRC-based model for SLR. As explained previously, ridge regression was applied to compute the output weight matrix (\( W_{\text{out}} \)) for label classification. Ridge regression extends standard linear regression by introducing a regularization term, which helps mitigate overfitting and enhances the model's ability to generalize to unseen data \cite{19,20}.

Ridge regression is a technique used to address some of the limitations of traditional linear regression, especially when dealing with multicollinearity among predictor variables. In conventional linear regression, the objective is to determine a coefficient vector \( \boldsymbol{\beta} \) that provides the best fit to the data by minimizing the squared error between the actual observed outcomes and the values predicted by the model. This method works well when the predictor variables are not highly correlated. However, when multicollinearity is present, the estimates of \( \beta \) can become unstable and exhibit high variance, leading to unreliable predictions.

To mitigate this issue, ridge regression modifies the linear regression objective function by adding a penalty term to the linear regression model. This penalty is proportional to the sum of the squares of the coefficients, effectively shrinking them towards zero. The modified objective function that ridge regression seeks to minimize is given below in Eq.~(\ref{eq:ridge_1}) \cite{21}.

\begin{equation}
L(\beta) = \sum_{i=1}^{n} (y_i - \mathbf{x}_i^\top \beta)^2 + \lambda \sum_{j=1}^{p} \beta_j^2
\label{eq:ridge_1}
\end{equation}

Here, \( y_i \) is the observed response for the \( i \)-th observation, \( \mathbf{x}_i \) is the vector of predictor variables for the \( i \)-th observation, \( \beta \) is the vector of coefficients, \( \lambda \) is a non-negative tuning parameter that controls the strength of the penalty, and \( p \) is the number of predictor variables. Each observation refers to a single data point consisting of a response \( y_i \) and a corresponding predictor vector \( \mathbf{x}_i \in \mathbb{R}^p \), where \( \mathbb{R}^p \) denotes a \( p \)-dimensional real-valued vector space. In other words, \( \mathbf{x}_i \) contains \( p \) real-valued features for the \( i \)-th sample. The index \( i = 1, \dots, n \) sums over all \( n \) observations in the dataset to compute the total prediction error, while the index \( j = 1, \dots, p \) sums over all \( p \) model coefficients to apply the regularization penalty on each feature weight. The first term in the equation is the residual sum of squares, which measures the model's fit to the data. The second term, \( \lambda \sum_{j=1}^{p} \beta_j^2 \), is the penalty term that discourages large coefficients. The solution to this optimization problem can be expressed in a closed form as written in Eq.~(\ref{eq:ridge_2}) \cite{22}.

\begin{equation}
\beta_{\text{ridge}} = (X^\top X + \lambda I)^{-1} X^\top y
\label{eq:ridge_2}
\end{equation}

Here, \( X \) is the matrix of predictor variables, \( X^\top \) is the transpose of \( X \), \( y \) is the vector of observed responses, and \( I \) is the identity matrix. The term \( \lambda I \) added to \( X^\top X \) ensures that the matrix is invertible, even when \( X^\top X \) is singular or nearly singular due to multicollinearity. By incorporating this penalty, ridge regression helps control the variance of coefficient estimates while introducing a slight bias. This balance improves the model’s stability and interpretability, particularly in cases where predictor variables are highly correlated or when the number of predictors surpasses the number of observations. Ridge regression enhances the robustness of linear regression models by adding a regularization term that penalizes significant coefficients, thereby addressing issues related to multicollinearity and overfitting \cite{23,24}. 

In the context of ridge regression applied to the output layer of a reservoir computing model, the output weight matrix \( W_{\text{out}} \) plays the same role as the coefficient vector \( \beta \) in standard linear regression. When training the output layer using ridge regression, the closed-form solution for \( W_{\text{out}} \) is given by Eq.~(\ref{eq:ridge_output_weights}):

\begin{equation}
W_{\text{out}} = \left( X^\top X + \lambda I \right)^{-1} X^\top Y
\label{eq:ridge_output_weights}
\end{equation}

Here, \( X \in \mathbb{R}^{T \times N} \) is the matrix of collected reservoir states (from time steps \( t=1 \) to \( T \)), \( N \) is the reservoir size, \( Y \in \mathbb{R}^{T \times C} \) is the matrix of target outputs or class labels, \( I \in \mathbb{R}^{N \times N} \) is the identity matrix used for regularization, and \( W_{\text{out}} \in \mathbb{R}^{C \times N} \) is the learned output weight matrix.

\subsection{Bidirectional gated recurrent unit (Bi-GRU)}

We also employed a deep learning-based method, Bi-GRU, for sign language recognition and compared the results with those of the PBRC-based architecture. We fed the keypoints extracted from the WLASL100 video dataset (using MediaPipe) to the Bi-GRU-based architecture. A Bi-GRU is an RNN-based architecture that processes sequential data in both forward and backward directions \cite{25}. The Bidirectional GRU (Bi-GRU) extends the standard GRU architecture, which itself is a simplified variant of LSTM. The internal operations of a GRU unit at a given time step \( t \) are described by Eqs.~(\ref{eq:gru_update})--(\ref{eq:gru_hidden}):

\begin{equation}
z_t = \sigma(W_z x_t + U_z h_{t-1} + b_z),
\label{eq:gru_update}
\end{equation}

\begin{equation}
r_t = \sigma(W_r x_t + U_r h_{t-1} + b_r),
\label{eq:gru_reset}
\end{equation}

\begin{equation}
\tilde{h}_t = \mathcal{A}(W_h x_t + U_h (r_t \odot h_{t-1}) + b_h),
\label{eq:gru_candidate}
\end{equation}

\begin{equation}
h_t = (1 - z_t) \odot h_{t-1} + z_t \odot \tilde{h}_t,
\label{eq:gru_hidden}
\end{equation}

where \( z_t \) and \( r_t \) denote the update and reset gate vectors, respectively, \( \tilde{h}_t \) is the candidate activation, \( h_t \) is the hidden state, and \( x_t \) is the input at time \( t \). The functions \( \sigma \) and \( \mathcal{A} \) represent the sigmoid and activation functions (commonly \( \tanh \)), and \( \odot \) indicates element-wise multiplication. The matrices \( W \), \( U \), and bias vector \( b \) are learnable parameters.

A Bi-GRU consists of two GRU layers: one processes the input sequence \( u(t) \) in the forward temporal direction (\( t = 1 \) to \( T \)) to generate forward hidden states \( \overrightarrow{h}_t \), while the other processes the sequence in reverse (\( t = T \) to \( 1 \)) to produce backward hidden states \( \overleftarrow{h}_t \). The final hidden representation at each time step is obtained by concatenating these two directional states, as given in Eq.~(\ref{eq:bigru_concat_oplus}):

\begin{equation}
h_t = \overrightarrow{h}_t \oplus \overleftarrow{h}_t,
\label{eq:bigru_concat_oplus}
\end{equation}

where \( \oplus \) denotes vector concatenation.

Each GRU node acts as a computational unit within the hidden layer, maintaining an internal state and performing gating operations at every time step. These nodes sequentially process the input, updating their hidden states based on both current and past information. A hidden layer in a GRU contains multiple such nodes operating in parallel. Although they share the same input, each node processes it independently, enabling diverse feature extraction. The number of nodes in the hidden layer determines the dimensionality of the hidden state vector \( h_t \). In Bi-GRU, both the forward \( \overrightarrow{h}_t \) and backward \( \overleftarrow{h}_t \) hidden states have this dimensionality, resulting in a final concatenated representation that doubles the hidden vector size.

\subsection{Experimental Settings}

 Here, we employed parallel bidirectional-based RC (70 nodes in each BRC), single bidirectional reservoir computing (140 nodes), deep learning-based methods such as Bi-GRU (150 epochs, nine hidden layers), and single standard ESN-based RC (unidirectional RC with 280 nodes) for SLR. We used MediaPipe to extract keypoints from the WLASL100 dataset in all methods. All experiments were performed on an Intel\textsuperscript{\textregistered}, Core\textsuperscript{™} i7 processor (32 GB of random access memory).

The hyperparameters used in our PBRC model are as follows: the spectral radius (\( \rho \)) is set to 0.3 for each BRC, the number of BRCs (\( N_{\mathrm{BRC}} \)) is 2, the number of reservoir nodes in each BRC (\( N_r \)) is 70, and the leak rate (\( \alpha \)) is 0.6 for each BRC.

The above-mentioned hyperparameters of the PBRC model were selected through hyperparameter tuning to optimize performance, and the final values were determined empirically based on the settings that achieved the highest validation accuracy. 

\subsection{Proposed method}

\begin{figure}[ht]
\centering
\includegraphics{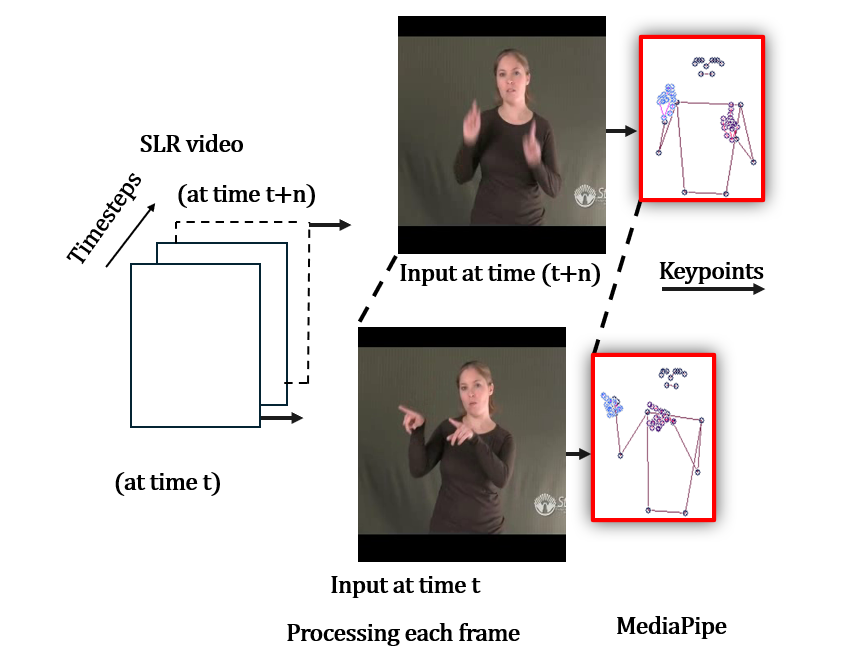}
\caption{Feature extraction using MediaPipe}
\label{fig:5}
\end{figure}

In this study (SLR using PBRC), we employed MediaPipe to extract skeletal keypoints from video frames, as shown in Fig.~\ref{fig:5}.

These keypoints represent specific anatomical landmarks on the hands of the sign language user, which are crucial for understanding the gestures. The keypoints are graphically depicted, showing the connections between different hand joints. 

\begin{figure}[t]
\centering
\includegraphics{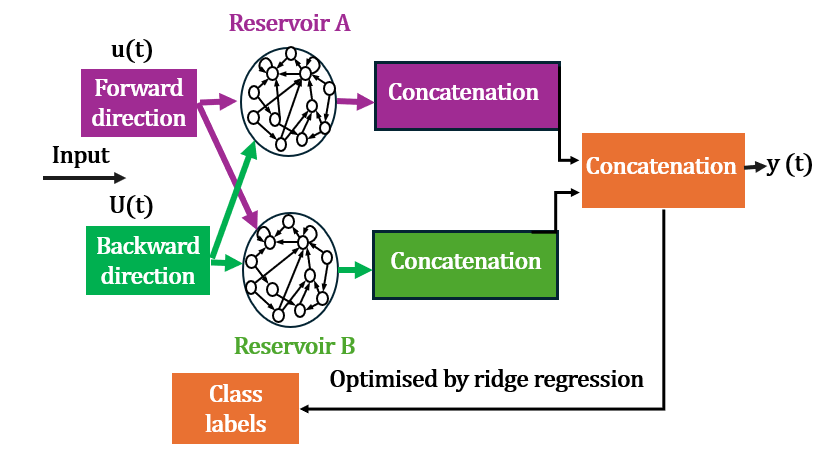}
\caption{Classification of SLR using PBRC}
\label{fig:6}
\end{figure}

The extracted keypoints are fed into the ESN-based PBRC architecture, which processes the input sequence \( u(t) \) and its reverse \( U(t) \) in parallel using two independently configured bidirectional reservoirs, as shown in Fig.~\ref{fig:6}. 

Each BRC extracts temporal features from both directions independently. Finally, the bidirectional outputs from both BRCs are concatenated to form a unified and comprehensive feature set that captures the full temporal dynamics of the gesture sequence. The final step involves classifying these features into specific class labels using a regression model, which represents the different signs in the WLASL100 dataset. Here, ridge regression is used to describe the bidirectional reservoir states and map these representations to the final output labels. The PBRC stage captures the most relevant patterns from the high-dimensional data, reducing noise while retaining meaningful information. This ensures that the learned representations are stable and informative. The second stage of ridge regression helps form the final decision boundaries by minimizing classification errors.

\subsection{Performance metrics}
We use accuracy to evaluate the performance of the PBRC-based SLR system. 
Accuracy measures the proportion of instances that the model classifies correctly out of all instances and is defined as shown in Eq.~(\ref{eq:accuracy}) \cite{26}:

\begin{equation}
\text{Accuracy} = \frac{\text{TP} + \text{TN}}{\text{TP} + \text{TN} + \text{FP} + \text{FN}}
\label{eq:accuracy}
\end{equation}

where $\text{TP}$ (true positives) is the number of correctly predicted positive class instances, $\text{TN}$ (true negatives) is the number of correctly predicted negative class instances, $\text{FP}$ (false positives) is the number of negative instances incorrectly predicted as positive, and $\text{FN}$ (false negatives) is the number of positive instances incorrectly predicted as negative. This metric indicates the model's ability to accurately classify input gestures into their respective classes.

\section{Results and discussion}

Table~\ref{tab:accuracy_baselines} compares the performance of four SLR methods implemented on the WLASL100 video dataset: the proposed PBRC-based approach, BRC, single unidirectional ESN, and Bi-GRU. In all the compared methods, we extracted keypoints from the sign video using MediaPipe and fed them into the corresponding reservoir architectures (i.e., standard single ESN, Bi-GRU, BRC, PBRC) for classifying sign language data. The metrics shown in Table~\ref{tab:accuracy_baselines} include accuracy (with standard deviation) and training time (average of five experiments).

\begin{table}[ht]
\centering
\caption{Comparison of SLR using PBRC (proposed method), BRC, Bi-GRU, and standard ESN on WLASL100}
\label{tab:accuracy_baselines}
\begin{tabular}{l c c c}
\hline
Method & Accuracy (\% $\pm$ SD) & Training time (mm:ss.ms) & Device \\
\hline
PBRC         & 60.85 $\pm$ 1.38 & 00:18.67 & CPU \\
BRC          & 58.11 $\pm$ 1.45 & 00:12.33 & CPU \\
Standard ESN & 56.90 $\pm$ 1.34 & 00:21.10 & CPU \\
Bi-GRU       & 50.01 $\pm$ 2.58 & 55:41.50 & CPU \\
\hline
\end{tabular}
\end{table}

As shown in Table~\ref{tab:accuracy_baselines}, the proposed PBRC-based approach achieves a competitive accuracy of \(60.85\%\) with a significantly lower training time of 18.67 seconds, outperforming the deep learning-based Bi-GRU. 

To demonstrate the potential of our model, we also calculated the Top-1, Top-5, and Top-10 accuracies of the proposed architecture, as shown in Table~\ref{tab:accuracy_comparison}.

\begin{table}[ht]
\centering
\caption{Accuracy comparison of PBRC with DL-based approaches (Pose-TGCN, I3D) and MRC for SLR on the WLASL100 dataset}
\label{tab:accuracy_comparison}
\begin{tabular}{l c c c c c c}
\hline
\text{Methods} & \multicolumn{3}{c}{\text{Accuracy (\%)}} & \text{Training time} & \text{Device} & \text{Reference} \\
\cline{2-4}
 & \text{Top-1} & \text{Top-5} & \text{Top-10} & (hh:mm:ss.ms) &  &  \\
\hline
PBRC        & 60.85 & 85.86 & 91.74 & 00:00:18.67 & CPU & Proposed method \\
Pose-TGCN   & 55.43 & 78.68 & 87.60 & 00:38:18.9  & GPU & ~\cite{27} \\
I3D         & 65.89 & 84.11 & 89.92 & 20:13:42.5  & GPU & ~\cite{27} \\
MRC         & 60.35 & 84.65 & 91.51 & 00:00:52.7  & CPU & ~\cite{27} \\
\hline
\end{tabular}
\end{table}

Table~\ref{tab:accuracy_comparison} shows the Top-1, Top-5, and Top-10 accuracy comparisons between the proposed PBRC model and existing deep learning-based approaches for sign language recognition on the WLASL100 dataset. The PBRC model achieves a Top-1 accuracy of \(60.85\%\), outperforming Pose-TGCN (\(55.43\%\)), while being competitive with the I3D model (\(65.89\%\)). We also calculated the training time for PBRC at 70 nodes, which was 18.67 seconds. Apart from accuracy, the proposed method significantly reduces training time compared to deep learning models such as Bi-GRU.

From the above, we can conclude that the proposed PBRC model demonstrates clear advantages over deep learning-based methods such as Bi-GRU and I3D. Compared to Bi-GRU, PBRC achieves higher accuracy while drastically reducing training time. Against I3D, PBRC delivers competitive Top-1, Top-5, and Top-10 accuracies while requiring substantially fewer computational resources, highlighting its efficiency for sign language recognition on edge devices.

The proposed architecture combines MediaPipe's efficient spatial keypoint extraction with the dynamic temporal modeling capabilities of PBRC to advance SLR. Unlike a conventional single bidirectional reservoir, the parallel configuration allows for independent forward and backward temporal processing across two distinct BRCs. This configuration enables the capture of richer and more complementary temporal features, leading to a more robust representation of gesture dynamics. Furthermore, since only the readout layer requires training via ridge regression, the overall training process remains lightweight and fast, without the need for extensive resources, making the system well-suited for real-time applications on resource-constrained edge devices.

\section{Conclusions}

This article presents a comparative analysis of the proposed PBRC method with the deep learning-based Bi-GRU model in terms of both accuracy and training time for SLR. Additionally, we compared our results with those of other approaches, such as I3D, Pose-TGCN, and MRC, applied to the WLASL100 dataset. The results demonstrate that the proposed PBRC model achieves Top-1, Top-5, and Top-10 accuracies of 60.85\%, 85.86\%, and 91.74\%, respectively, showing competitive performance compared to other deep learning-based SLR systems. Apart from accuracy, the proposed method significantly reduces training time to 18.67 seconds due to the intrinsic properties of RC, compared to more than 55 minutes for deep learning-based methods like Bi-GRU, showing its potential for deployment on edge devices. 

While the proposed PBRC-based approach demonstrates competitive accuracy for isolated sign language recognition, there is still scope for improvement. Increasing recognition accuracy remains a challenging task, and future work should focus on exploring optimized reservoir configurations to enhance performance. Currently, the system is limited to isolated sign recognition, where each sign (word) is recognized independently. This PBRC-based SLR framework can be extended to continuous sign language recognition, where signs are identified seamlessly within natural sentence structures.

\vspace{0.5em}
\noindent\textbf{Acknowledgements:}
This work was financially supported in part by the New Energy and Industrial Technology Development Organization (NEDO), the Japan Science and Technology Agency (JST), and the Japan Society for the Promotion of Science (JSPS). Specifically, this research was supported by NEDO, Japan (Grant No. JPNP16007); JST ALCA-Next, Japan (Grant No. JPMJAN23F3); and JSPS KAKENHI, Japan (Grant Nos. 22K17968, 23H03468, and 23K18495).

\vspace{0.5em}

\noindent\textbf{Conflict of Interest:}
The authors declare no conflict of interests.
\vspace{0.5em}

\noindent\textbf{Author Contributions:} Conceptualization: NKS. Funding acquisition: YT, HT. Investigation: NKS. Supervision: YT, HT. Visualization: NKS. Writing--original NKS. Writing--review \& editing: ARS, YT, HT.

\section*{Appendix: Links}

\begin{itemize}

  \item[\textbullet] Copyright 2026 IEICE, This article has been accepted by Nonlinear Theory and Its Applications, IEICE (NOLTA):Vol.E17-N,No.1,pp.-,Jan. 2026\\
  \url{https://www.jstage.jst.go.jp/browse/nolta/-char/en}

  \item[\textbullet] Video abstract:  
  \url{https://www.youtube.com/watch?v=iDilgLZ7gCA}

\item[\textbullet] WLASL video dataset: \url{https://dxli94.github.io/WLASL/} 
\end{itemize}

\end{document}